\documentclass[runningheads]{llncs}
\usepackage[T1]{fontenc}
\usepackage{graphicx}
\usepackage{booktabs}
\usepackage{multirow}
\usepackage{subcaption}
\usepackage{amssymb}
\usepackage[misc]{ifsym}

\usepackage{mwe}

\begin{document}

\title{GAPNet: Plug-in Jointly Learning Task-Specific Graph for Dynamic Stock Relation}

\titlerunning{GAPNet}

\author{Yingjie Niu\inst{1}\orcidID{0000-0001-9322-2726} \and
Lanxin Lu\inst{1}\orcidID{0000-0003-1094-0517} \and
Changhong Jin\inst{1}\orcidID{0000-0003-2565-592X} \and 
Ruihai Dong\inst{1}\orcidID{0000-0002-2509-1370}}

\authorrunning{Niu et al.}

\institute{University College Dublin, Dublin, Ireland}

\maketitle              

\begin{abstract}
    The advent of the web has led to a paradigm shift in the financial relations, with the real-time dissemination of news, social discourse, and financial filings contributing significantly to the reshaping of financial forecasting. 
    The existing methods rely on establishing relations a priori, i.e. predefining graphs to capture inter-stock relationships.
    However, the stock-related web signals are characterised by high levels of noise, asynchrony, and challenging to obtain, resulting in poor generalisability and non-alignment between the predefined graphs and the downstream tasks. 
    To address this, we propose \textbf{GAPNet}, a \textbf{G}raph \textbf{A}daptation \textbf{P}lug-in \textbf{Net}work that jointly learns task-specific topology and representations in an end-to-end manner. 
    GAPNet attaches to existing pairwise graph or hypergraph backbone models, enabling the dynamic adaptation and rewiring of edge topologies via two complementary components: a Spatial Perception Layer that captures short-term co-movements across assets, and a Temporal Perception Layer that maintains long-term dependency under distribution shift. 
    Across two real-world stock datasets, GAPNet has been shown to consistently enhance the profitability and stability in comparision to the state-of-the-art models, yielding annualised cumulative returns of up to 0.47 for RT-GCN and 0.63 for CI-STHPAN, with peak Sharpe Ratio of 2.20 and 2.12 respectively. 
    The plug-and-play design of GAPNet ensures its broad applicability to diverse GNN-based architectures. 
    Our results underscore that jointly learning graph structures and representations is essential for task-specific relational modeling.

\end{abstract}

\section{Introduction}
Globally, the daily volume of transactions in stock markets is measured in billions of dollars. 
The stock market is regarded as one of the most substantial and liquid components of the global financial system. 
It plays a pivotal role in every growing and thriving economy, with the total value of stocks traded reaching $117.7\%$ of GDP in 2024. 
The stock market is characterised by inherently volatile and exhibit complex temporal and cross-asset dependencies, resulting in the longstanding challenge of accurate forecasting. 
A number of studies modeled stock prices as univariate time series, prompting the adoption of recurrent neural networks (RNNs), including long short-term memory (LSTM) networks~\cite{feng2018enhancing}, gated recurrent units (GRUs)~\cite{xu2018stock}, and Transformer-based methods~\cite{muhammad2023transformer}. 
However, such sequential models typically overlook the inter-stock relationships.

The web has become the beating heart of financial information flow, with breaking news, social discourse and financial filings spreading online, reshaping financial forecasting in real time.
As these web signals induce contextual, time-varying relations among companies, a growing number of studies employ graph neural networks (GNNs) to aggregate information across related stocks by explicitly modeling their relationships. 

\begin{figure}[t]
\centering
\includegraphics[width=\linewidth]{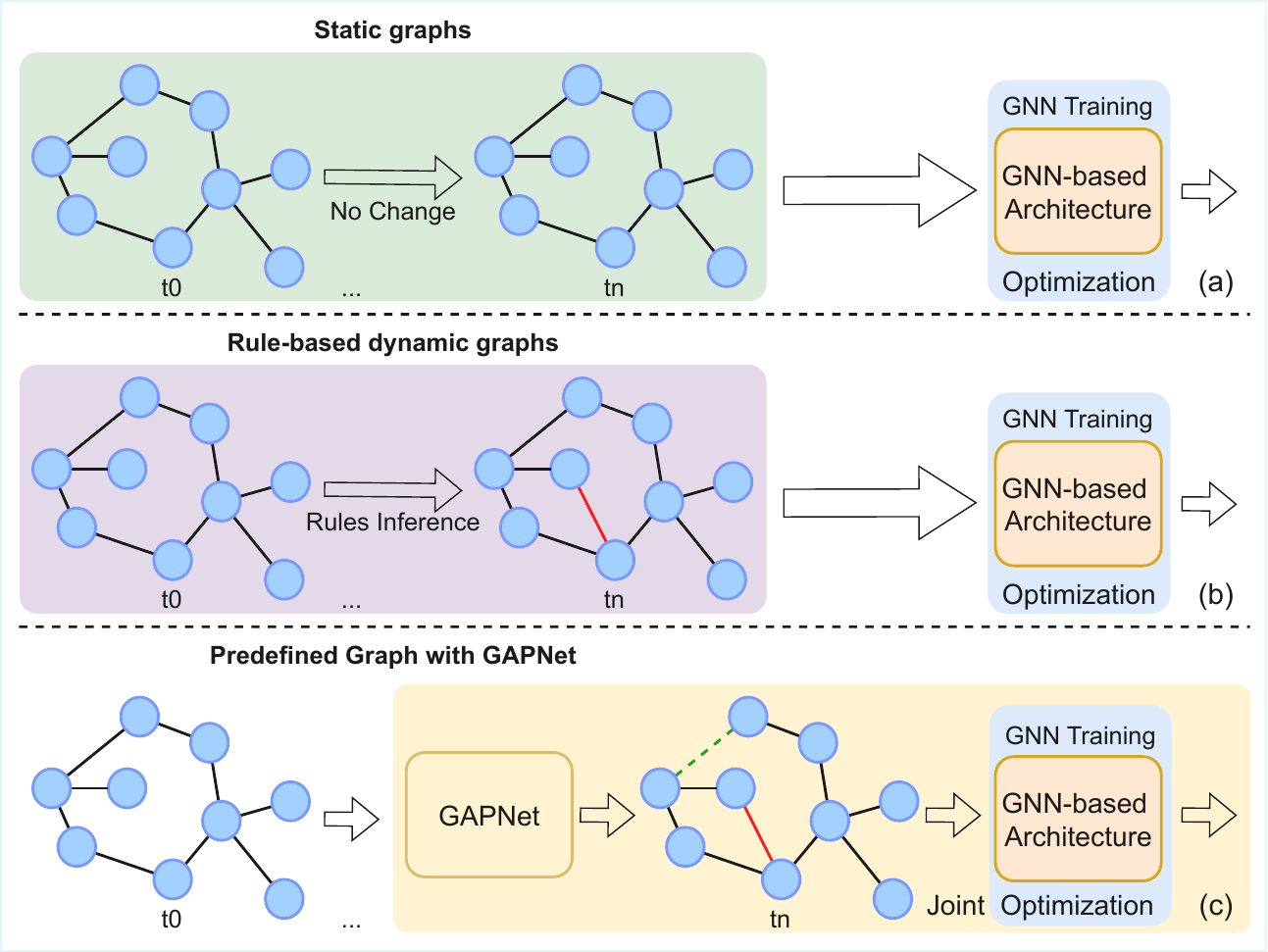} 
\caption{Training paradigm comparison from $t0$ to $tn$. (a) Two-step training paradigm using static graphs; (b) Two-step training paradigm using rule-based dynamic graphs; (c) End-to-end training paradigm using GAPNet.}
\label{fig:paradigm}
\end{figure}

The methods typically follow a two-step training paradigm with the use of static graphs or rule-based dynamic graphs, which can be simplified into \textbf{graph construction $\rightarrow$ GNN training}, are shown in Figure~\ref{fig:paradigm}a and Figure~\ref{fig:paradigm}b. 
The \textbf{static graphs} based on long-lasting relationships, e.g., sector information~\cite{feng2019temporal,cui2023temporal}, and momentum spillover effects~\cite{cheng2021modeling} remain unchanged.
The \textbf{rule-based dynamic graphs} where edges are constructed via rules inference, e.g., correlation thresholds~\cite{xiang2022temporal}, news co-occurrence~\cite{niu2024evaluating,niu2025ngat}, and dynamic time warping (DTW)~\cite{sawhney2021stock,xia2024ci}, evolve over time while remaining separate from the process of GNN training.
Consequently, this two-step training paradigm limits generalisability and often fails to align with the downstream task, due to the graph being constructed independently before the optimisation process. 

Graph structures, defined from a variety of sources, possess their own distinctive benefits and biases. 
Some studies employ more complicated rules and involve multiple data sources to reduce bias. 
However, such approaches generally increase the complexity of the graph construction process (e.g. spatial-temporal graph~\cite{zheng2023relational}) or rely on information that is difficult to obtain (e.g. shared holdings~\cite{chen2018incorporating}), thereby further limiting the generalisability of the models. 
In the context of predefined graph structures limitations and the difficulty in obtaining stock relationship data, the question of how to use readily available, high-frequency stock price information to reduce bias while jointly learning downstream tasks remains unresolved. 
In this study, the exploration can be defined as an alignment operation between a predefined graph structure and a downstream task using stock price information.

Beyond the constraints introduced by the training paradigm, the predefined graph construction approaches themselves face two critical issues that remain to be addressed: 
\begin{enumerate}

\item \textbf{Limited node receptive field:} the node receptive field is defined as the number of nodes considered when establishing an edge, which is inherently constrained to pairwise in present rule-based dynamic graphs (e.g., the similarity between stocks $s_i$ and $s_j$ in DTW). Pairwise node receptive field overlooks potentially crucial higher-order relationships within stock clusters which is shown to be more suitable than pairwise similarity in financial forecasting tasks~\cite{sawhney2021stock,duan2025factorgcl}. This underscores the importance of considering a broader node receptive field during edge construction; 
\item \textbf{Single focus:} static graphs are usually built on long-term relationships while the rule-based dynamic ones tend to capture only short-term dependencies. For example, rolling correlation edges on day t depend solely on the preceding rolling window, ignoring historical context, while the industry graph only considers the stock industry information which exists for long term. 
\end{enumerate}

To address the above issues, we propose a novel \textbf{G}raph \textbf{A}daptation \textbf{P}lug-in \textbf{Net}work (\textbf{GAPNet}) and achieve an end‑to‑end training paradigm, which is \textbf{graph construction $\rightarrow$ GAPNet $\rightarrow$ GNN training} are shown in Figure \ref{fig:paradigm}c. 
The paradigm enables graph structure alignment and GNN training to proceed simultaneously by plugging in a GAPNet between the graph construction and GNN architectures. 
Specifically, GAPNet dynamically refines the graph structure in a task-oriented manner, aligning it with the downstream task. 
Accordingly, it provides GNN models with the capability to adapt any given predefined graph by removing task-irrelevant or even harmful biases and injecting informative relations that enhance task performance. 
GAPNet also addresses the limited generalisation problem of the traditional two-step training paradigm. 
Starting from any graph structure, GAPNet continually updates the graph structure with stock price information to align the graph structure more effectively with downstream tasks, which enables the model to achieve considerably enhanced performance on tasks. 
This has the effect of reducing the impact of graph complexity and the difficulty to obtain information on model generalisation. 
Furthermore, the issues with existing graph construction approaches are addressed by the GAPNet architecture: (1) the \textit{Spatial Perception Layer} that expands the node receptive field beyond pairwise by involving attention from a broader set of candidate nodes; and (2) the \textit{Temporal Perception Layer} that incorporates both short-term and long-term memory into graph construction. Our main contributions include:
\begin{itemize}
  \item Propose a novel end-to-end training paradigm for GNN-based financial forecasting, resolving the non-alignment between predefined graph structures and the downstream tasks and improving generalisability.
  \item Introduce GAPNet, a modular plug-in network between predefined graphs and GNN architectures to refine the graph in a task-oriented manner. 
  \item Address the limited node receptive and single focus problems in graph construction approaches by designing the spatial and temporal perception layers. 
  \item Extensive experiments on two real‑world stock datasets (i.e., NASDAQ and NYSE) demonstrate consistent and statistically significant improvements when integrating GAPNet with multiple SOTA backbone models.
\end{itemize}

\begin{figure*}[t]
\centering
\includegraphics[width=\textwidth]{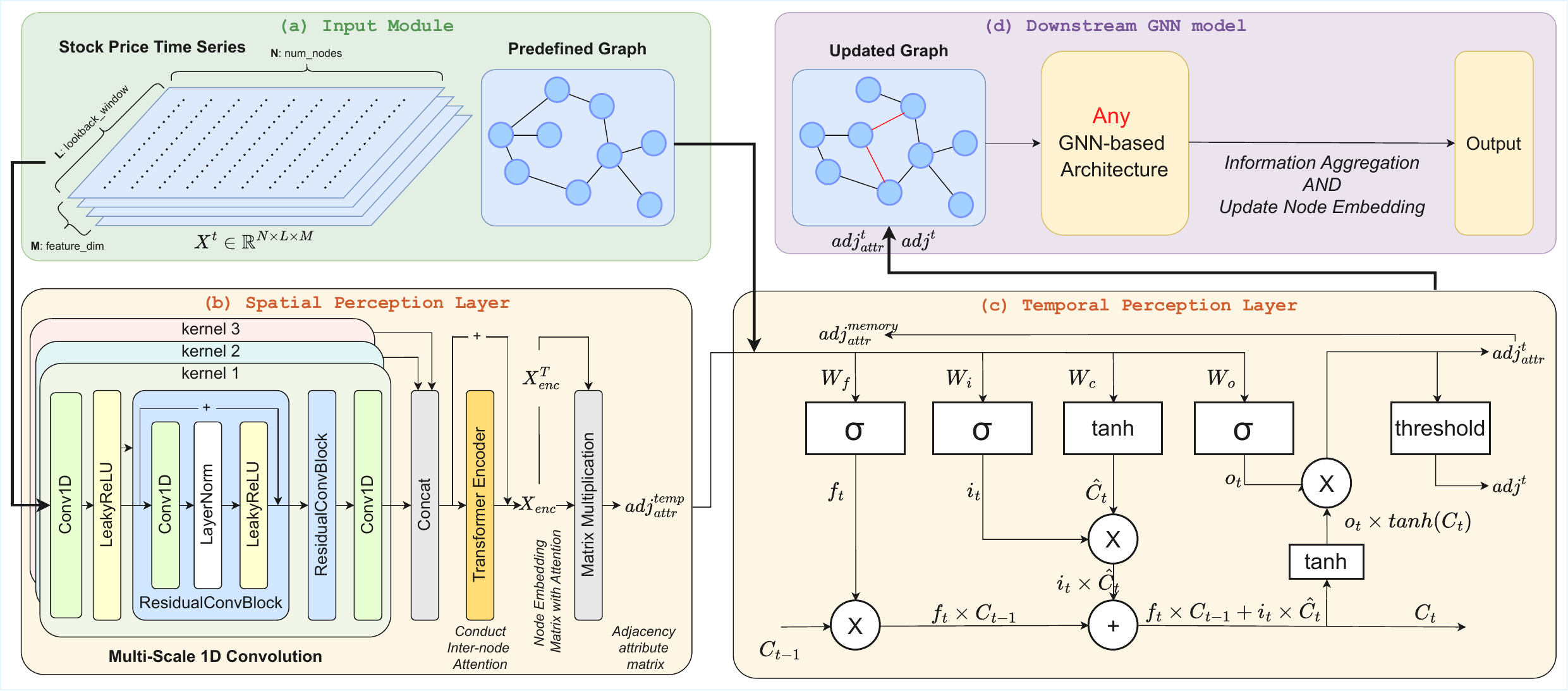} 
\caption{Model architecture of GAPNet and an overview of the end-to-end training paradigm on the stock ranking task.}
\label{fig:model-architecture}
\end{figure*}



\section{Related Work}
\subsection{Pairwise Graph-based Methods}
Graph neural networks (GNN)~\cite{kipf2016semi,velivckovic2017graph} have become a leading paradigm for modeling inter-stock relationships~\cite{patel2024systematic}. 
Early studies typically constructs stock-relation graphs intuitively and then applied traditional GNNs. 
For instance, \cite{chen2018incorporating} derive a firm investment graph and proposes a stock prediction model that that aggregates signals from related stocks. 
\cite{feng2019temporal} propose RSR-I and build industry co-membership and Wikipedia-linkage graphs to introduce domain knowledge into the temporal relation model. 
Furthermore, \cite{xu2022hgnn} propose the use of hierarchical (stock-industry-market) graphs to predict price-limits. 
However, static, predefined graphs often fail to reflect the rapidly evolving and dynamic dependencies in the financial market. 
In response, recent studies have increasingly enhance the graph construction and GNN architectures to improve performance on stock forecasting. 
For instance, \cite{li2021modeling} build construct correlation-based stock graphs and propose an LSTM-RGCN to predict the overnight stock movement, demonstrating the benefit of incorporating cross-asset relations beyond single-stock signals. 
\cite{cheng2021modeling} introduce AD-GAT, which models momentum spillovers by weighting graph edges with firm attributes. 
\cite{xiang2022temporal} compute the dynamic rolling correlation, and develop temporal and heterogeneous graph neural network-based (THGNN) approach to learn the dynamic relations. 
\cite{zheng2023relational} design relational temporal convolutional networks (RT-GCN) to capture spatial relations and temporal dynamics for stock ranking prediction. \cite{qian2024mdgnn} present a multi-relational dynamic (MDGNN) framework that builds the discrete dynamic graph to capture multifaceted relations. 

\subsection{Hypergraph-based Methods}
Compared with pairwise graphs, hypergraphs provide the capability to encode high-order relations among multiple entities~\cite{zhou2006learning}. 
Motivated by this advantage, STHAN-SR~\cite{sawhney2021stock} has constructed financial relational hypergraphs to express higher-order relations that exist in stock clusters. 
ALSP-TF~\cite{wang2022adaptive} and CI-STHPAN~\cite{xia2024ci} have evolved further from explicitly defined relations to data-driven hypergraph similarity adjacency modeling. 
Beyond those relations, \cite{duan2025factorgcl}  exploits hypergraph structures to model high-order interactions between stock returns and factors.

\subsection{Graph Learning}
A growing line of studies has emerged that focuses on learnable graph construction, jointly learning the graph structure with the predictor.
\cite{jiang2019semi} introduce GLCN, which integrates a learnable pairwise adjacency matrix into GCN training for semi-supervised classification. 
Additionally, \cite{zhang2020spatio} propose structure learning convolution to infer both global and local graph structures for spatio-temporal traffic forecasting. 
In the context of time-series settings, \cite{shang2021discrete} learn discrete graph structures end-to-end to enhance multi-series forecasting.
For graph learning itself, \cite{franceschi2019learning} and \cite{chen2020iterative} propose bilevel/iterative frameworks to jointly optimise topology and GNN parameters.
In the field of financial forecasting, \cite{you2024dgdnn} propose DGDNN for constructing dynamic stock graphs via entropy-based edge generation, subsequently employing diffusion-based representation learning for stock trend classification.

\section{GAPNet Design}
\subsection{Problem Framework}
Following previous studies on graph- and hypergraph-based financial modeling~\cite{feng2019temporal,sawhney2021stock,wang2022adaptive,xia2024ci}, we consider stock ranking problem as our downstream task, which selects stocks that maximise investment profits. 
Specifically, let $S = \{s_1, s_2, ..., s_N\}$ denote the set of $N$ stocks. 
For each stock $s_i \in S$, let $p_i^t$ denote the closing price on dat $t$ and let $r_i^t$ denote the one-day return ratio, defined as $r_i^t = \frac{p_i^t - p_i^{t-1}}{p_i^{t-1}}$. 

The following rule is to be employed to establish a ranking of the stocks: \textit{for any two stocks $s_i, s_j \in S$, if $r_i^t>r_j^t$, then it can be deduced that $y_i^t>y_j^t$ where $y_i^t$ is the ranking score of stock $s_i$ on trading day $t$}. 
On any given trading day $t$, there exists an optimal ranking, denote by $Y^t = \{y_1^t > y_2^t>...>y_N^t\}$, where the higher the ranking of a stock, the higher the expected profit. 
In the context of stock trading data for a lookback window of size $L$ (i.e. $X_i^t = [x_i^{t-L},...,x_i^{t-1}] \in \mathbb{R}^{L \times M}$ where $M$ denotes the feature dimension, and $\mathbf{X}^t = [X_1^t, X_2^t, ..., X_N^t] \in \mathbb{R}^{N \times L \times M}$), we aim to develop a ranking algorithm that can output the ranking list of stocks $S$. The top-ranked $k$ stocks are then selected for investment on day $t$.

Since the selected downstream task is stock ranking, a task-specific loss function combining point-wise regression loss and pairwise ranking loss is applied. 
\begin{equation}
    \mathcal{L} = ||\hat{r}_i^t - r_i^t||^2 + \alpha \sum_{i=1}^{|S|} \sum_{j=1}^{|S|} max(0 - (\hat{r}_i^t - \hat{r}_j^t)(r_i^t - r_j^t))
\end{equation}
where $\alpha$ is a hyperparameter to balance the two loss terms.
Working under the end-to-end training paradigm, GAPNet updates the graph structures to align the downstream task. It can be replaced with any task-specific loss function depending on the downstream task selection.

\subsection{Model Architecture}
Figure~\ref{fig:model-architecture} demonstrates the GAPNet architecture and the end-to-end training paradigm under the stock ranking task, which comprises four components: (a) the input module, (b) the Spatial Perception Layer (SPL), (c) the Temporal Perception Layer (TPL), and (d) the downstream GNN model. 
Specifically, the stock price time series $X^t \in \mathbb{R}^{N \times L \times M}$ is first fed into the SPL to encode multi-scale temporal patterns and inter-node interactions (via residual 1D convolutions and attention). 
The SPL is responsible for the construction of a temporary adjacency attribute matrix $adj_{attr}^{t,temp}$, combined with the predefined stock-relation graph, which is then used as input by the TPL. 
The TPL employs gated temporal aggregation to refine edge attributes across time and outputs the final adjacency attribute matrix $adj_{attr}^t$, which yields an updated, task-aligned graph for day $t$.
Finally, any GNN-based architecture can use this updated graph together with node features to aggregate information and produce ranking scores for the stock-ranking task.

\subsubsection{Spatial Perception Layer (SPL)}
As demonstrated in the Figure \ref{fig:model-architecture}b, the SPL uses short-term information embedded in the input feature matrix $\mathbf{X}^t$ to construct a temporary adjacency attribute matrix $adj_{attr}^{temp}$. 
Initially, the input matrix is processed by multiple 1D convolutional modules with different scales to generate node-specific embeddings. Since GAPNet aims to be integrated as a plug-in module into any GNN architecture, a common challenge arises when it is appended to already deep and complex models. This integration has the potential to result in excessively deep networks, making GAPNet susceptible to vanishing gradients. Such gradients hinders effective parameter updates. 
In order to address this issue, we introduce a \textit{Residual Convolution Block} to facilitate gradient flow and ensure stable training.
\begin{equation}
    ResConv(x) = x + LeakeyReLU(Norm(Conv1D(x)))
\end{equation}
The convolution of each scale is comprised of two normal \textit{Conv1D} modules and two \textit{Residual Convolution Blocks}. These convolutions operate along the temporal dimension, where $K$ different sizes of kernels are employed to capture localized temporal patterns within varying time windows for each node. 
Subsequently, the output of each kernel is concatenated, representing the feature embedding of each node. 
In particular,  
\begin{equation}
    X_{conv} =||_{k=1}^K MultiScale1DConv_{k}(\mathbf{X}^t)
\end{equation}
where $X_{conv} \in \mathbb{R}^{N \times Z \times (K*L)}$, and $Z$ is the number of kernels of each size. Then, a Transformer Encoder \cite{vaswani2017attention} is employed to conduct inter-node self-attention and generate the new node representation with attention on all other nodes. 
\begin{equation}
    X_{enc} = ||_{h=1}^{H} [X_{conv}+TransformerEncoder(X_{conv})]
\end{equation}
where $H$ is the number of heads in the Transformer Encoder, and $X_{enc} \in \mathbb{R}^{Z \times N \times (H*K*L)}$. We then conduct a dot product between each pair of node embeddings to obtain the temporary attribute of the potential edge between these two nodes. 
\begin{equation}
    adj_{attr}^{temp} = X_{enc}X_{enc}^T  
\end{equation}
where $adj_{attr}^{temp} \in \mathbb{R}^{Z \times N \times N}$. 
For each pair of nodes, a vector of size $[Z,1]$ is used to represent the attribute of the potential edge between them. 
It is evident that, given each node representation in $X_{enc}$ is computed by attending to all other nodes, the edge attributes in $adj_{attr}^{temp}$ inherently possess a node receptive field of $N$.

\subsubsection{Temporal Perception Layer (TPL)}
The construction of edges based solely on SPL with a limited lookback window captures only short-term dynamics while ignoring long-term dependencies.  
The TPL structure, has thus been designed with reference to Long Short-Term Memory (LSTM) \cite{hochreiter1997long} networks. 
In this structure, the temporary adjacency attribute matrix $adj_{attr}^{temp}$ serves as the input of the current time step. 
In addition, at each time step $t$, two auxiliary inputs are maintained: a memory adjacency matrix $adj_{attr}^{memory}$ and a cell state $C_{t-1}$ to capture long-term structural dependencies. 
The function of $adj_{attr}^{memory}$ is to store the memory of the adjacency attribute matrix from the previous step, while $C_{t-1}$ is employed to store the global long-term memory of the adjacency attribute matrix. 

In our experiment, the TPL maintains a memory state that evloves across days. 
We initialise the cell state $C_{t-1}$ to zero and set the memory adjacency $adj_{attr}^{memory}$ from the predefined stock-relation graph (e.g. Industry/Wiki). 
As shown in Figure~\ref{fig:model-architecture}c, the calculation flow of TPL can be described as follows: the forget gate is responsible for determining the information to forget from the previous step.
\begin{equation}
    f_t = \sigma(W_f[adj_{attr}^{temp} || adj_{attr}^{memory}] + b_f)
\end{equation} 
The input gate controls the information from the current input and previous memory into the memory cell. 
\begin{equation}
    i_t = \sigma(W_i[adj_{attr}^{temp} || adj_{attr}^{memory}] + b_i)
\end{equation}
\begin{equation}
    \hat{C_t} = tanh(W_c[adj_{attr}^{temp} || adj_{attr}^{memory}] + b_c)
\end{equation}
The cell state is updated based on the forget gate and the input gate. 
\begin{equation}
    C_t = f_t \times C_{t-1} + i_t \times \hat{C_t}
\end{equation}
The output gate controls the information from the memory cell to the output at the current time step, which is denoted by $adj_{attr}^t$. 
\begin{equation}
    o_t = \sigma(W_o[adj_{attr}^{temp} || adj_{attr}^{memory}] + b_o)
\end{equation}
\begin{equation}
    adj_{attr}^t = o_t \times tanh(C_t)
\end{equation}
where $W_f, W_i, W_c, W_o \in \mathbb{R}^{N \times 2N}$ and $b_f, b_i, b_c, b_o \in \mathbb{R}^N$. $adj_{attr}^t \in \mathbb{R}^{Z \times N \times N}$ is the final adjacency attribute matrix on day $t$, which has considered both short-term trend $adj_{attr}^{temp}$ and long-term memories $adj_{attr}^{memory}$. 
$adj_{attr}^t$ will inherently become the long-term memory at time stamp $t+1$. 

It is worth noting that GAPNet remains operational when no predefined stock-relation graph is available.
In this case, the memory adjacency $adj_{attr}^{memory}$ is randomly initialised, while the remainder of the calculation flow remains unchanged. 

In practice, $adj_{attr}^t$ may assign non-zero attributes to any number of node pairs, which can result in an overly dense (and noisy) graph.
In order to avoid this, we have opted to construct $adj^t$ by thresholding. Specifically, an edge is only created when the average magnitude of the corresponding attribute vector in $adj_{attr}^t$ exceeds a threshold $\tau$ (i.e. $\vert{mean(adj_{attr}^t[:,s_i,s_j])\vert} > \tau$).

\begin{table*}[th!]
\centering
\resizebox{\linewidth}{!}{%
\begin{tabular}{lccrccccccc}
\hline
\multicolumn{1}{c}{\multirow{2}{*}{Dataset}} & \multicolumn{1}{c}{\multirow{2}{*}{Start date}} & 
\multicolumn{1}{c}{\multirow{2}{*}{End date}} & \multicolumn{1}{c}{\multirow{2}{*}{\# Stocks}} & \multirow{2}{*}{\begin{tabular}[c]{@{}c@{}}Trading\\  Data\end{tabular}} & \multicolumn{4}{c}{Stock-relation Data} & \multirow{2}{*}{\begin{tabular}[c]{@{}c@{}}Public \\ Available\end{tabular}} \\
\multicolumn{1}{c}{} & \multicolumn{1}{c}{} & \multicolumn{1}{c}{} & \multicolumn{1}{c}{} & & Industry & Wiki & DTW & Other relation & \\ 
\hline
ACL18~\cite{xu2018stock} & 01/02/2014 & 12/30/2015 & 87 & \checkmark & - & - & - & - & \checkmark \\
CIKM18~\cite{wu2018hybrid} & 01/03/2017 & 12/28/2017 & 38 & \checkmark & - & - & -& -& \checkmark \\
NASDAQ~\cite{feng2019temporal,xia2024ci} & 01/02/2013 & 1208/2017 & 1026 & \checkmark & \checkmark & \checkmark & \checkmark & - & \checkmark \\
NYSE~\cite{feng2019temporal,xia2024ci} & 01/02/2013 & 12/08/2017 & 1737 & \checkmark & \checkmark & \checkmark & \checkmark & - & \checkmark \\
SSE~\cite{you2024multi} & 01/2015 & 12/2019 & 130 & \checkmark & \checkmark & - & - & - & \checkmark \\
BigData22~\cite{soun2022accurate} & 07/05/2019 & 06/30/2020 & 50 & \checkmark & - & - & - & - & \checkmark \\
FNSPID~\cite{dong2024fnspid} & 1999 & 2023 & 4775 & \checkmark & - & - & - & - & \checkmark \\
CSI~\cite{qian2024mdgnn} & 01/01/2020 & 02/31/2023 & 400 & \checkmark & \checkmark & - & - & Bank investment & - \\
SPNews~\cite{niu2024evaluating} & 05/31/2021 & 05/21/2023 & 431 & \checkmark & - & - & - & News co-occurence & \checkmark \\ 
\hline
\end{tabular}%
}
\caption{The comparison of candidate datasets}
\label{tab:dataset_selection}
\end{table*}

Finally, we apply $adj^t$ as a mask to $adj_{attr}^t$ matrix, retaining only the attribute information of the selected edges while filtering out spurious connections. 
The final output of the GAPNet module: $adj^t$ and $adj_{attr}^t$, which respectively represent the connectivity between nodes and the learned embeddings for each edge, respectively. 
When combined with the original node feature representations from the input data, this forms the constructed graph. 

\subsubsection{Adjacency Matrix to Hypergraph}
The GAPNet is capable of supporting both standard pairwise graph alignment and hypergraph alignment. 
As previously described, for standard graphs, we produce an adjacency matrix with shape $[N,N]$ is constructed, along with its corresponding edge attribute matrix. 
In the case of hypergraphs, we similarly generate a matrix $hyperadj^t \in \mathbb{R}^{N \times N}$ and its associated attribute matrix. 
However, the semantics of $hyperadj^t$ differ from $adj^t$.
Indeed, the $i$-th row in $hyperadj^t$ corresponds to the $i$-th hyperedge, and $hyperadj^t[i,j]$ indicates whether node $j$ participates in hyperedge $i$. 
A row with all-zero entries signifies a non-existent hyperedge. In our experiments, we assume that the total number of hyperedges is less than or equal to the number of nodes $N$, as reflected by the fixed shape of the $hyperadj^t$ matrix, which is of size  $[N,N]$.

\section{Experiments}

\subsection{Dataset Selection}


Table~\ref{tab:dataset_selection} outlines the key features of the candidate datasets that significantly impact backtesting and graph construction. 
These features include temporal coverage (start/end dates), scale (number of stocks), the availability of stock-relation data (Industry, Wiki, DTW similarity and other relations) and public release status.
From dataset comparison, the following observations can be deduced: 
\begin{enumerate}
    \item \textbf{Temporal coverage:} the longest span of consecutive years is FNSPID (1999–2023), followed by NASDAQ, NYSE and SSE. However, CIKM18, BigData22 are limited in scope. 
    \item \textbf{Scale:} compared to ACL18, CIKM18, SSE, BigData22 (with a maximum of 130 stocks), NASDAQ, NYSE, and FNSPID provide significantly larger range of stocks, which is a crucial aspect for both stock ranking and portfolio construction.
    \item \textbf{Relational richness:} both NASDAQ and NYSE include multiple stock-relation data, while other candidate datasets provide only one or no relation type.
    \item \textbf{Openness:} CSI is not released, whereas other candidate datasets are publicly available, allowing for replication.
\end{enumerate}

Accordingly, we adopt NASDAQ and NYSE, which contain stock trading data from 2013 to 2017 and three types of stock relation data. We strictly follow the sequential order to split the dataset into training/validation/testing sets. 
Detailed statistic of the dataset and splitting is shown in Table \ref{tab:dataset}. 
Similar to previous works~\cite{feng2019temporal,xia2024ci}, we compute the 5-, 10-, 20- and 30-day moving averages of the closing price to represent the time series trend in stock trading data, yielding an input feature dimension $M = 5$. 
We then standardise the entire dataset based on the maximum value of each feature on the training set. 
For the graph data, we use the hypergraphs constructed by~\cite{xia2024ci} with different relations, including Industry, Wiki, DTW similarity and their combinations. 
Since we also experiment with models that use traditional pairwise graphs instead of hypergraphs, we transfer the hypergraphs to its corresponding pairwise graphs with the following rule: 

\textit{For any two nodes $s_i, s_j$ on the same hyperedge, we build a pairwise edge between $s_i, s_j$}.

\begin{table}[t]
\centering
\begin{tabular}{lcc}
\hline
\multicolumn{1}{c}{}   & NASDAQ                & NYSE                \\ \hline
Stocks(Nodes)          & 1,026                 & 1,737               \\
Training Period        & \multicolumn{2}{c}{01/02/2013 - 12/31/2015} \\
Validation Period      & \multicolumn{2}{c}{01/04/2016 - 12/30/2016} \\
Testing Period         & \multicolumn{2}{c}{01/03/2017 - 12/08/2017} \\
Days(Train:Valid:Test) & \multicolumn{2}{c}{756 : 252 : 237}         \\ \hline
\end{tabular}%
\caption{NASDAQ and NYSE dataset statistics}
\label{tab:dataset}
\end{table}

\subsection{Implementation Details}
Our experiments are implemented with PyTorch and PyTorch Geometric, and we fix the random seed as 2023. We align the lookback window length with the original settings of each backbone model, which are 336 for the CI–STHPAN model and 16 for all others. 
The convolutional kernels sizes $K \in \{3, 5, 7\}$, and the number of each size of kernel $Z \in \{1, 4\}$. The hidden units of SPL is selected from $\{16,32,64,128,256\}$. 
For the Transformer encoder, we use single-head attention, a feedforward dimension of 128, a dropout rate of 0.1, and one encoder layer. 
The threshold $\tau=0.5$. In all backtesting experiments, we construct a portfolio from the $top-5$ ranked stocks.
All our experiments are conducted on a GeForce RTX 4090 GPU with Adam optimiser~\cite{kingma2014adam}. The learning rate is dynamically adjusted using OnceCycleLR~\cite{smith2019super} with a maximum value of $1e-4$, and the loss weight is set to a value $\alpha \in \{0.1,0.2,1,2,4,6,8,10\}$. 
The number of epochs is selected from $\{50,100\}$ with the implementation of early stopping strategies.

\begin{table*}[]
\centering
\resizebox{\textwidth}{!}{%
\begin{tabular}{cllcc|cc}
\hline
\multirow{2}{*}{Dataset} & \multirow{2}{*}{Model Architecture} & \multirow{2}{*}{Graph Type} & \multicolumn{2}{c|}{Basic with DTW Graph} & \multicolumn{2}{c}{\textbf{GAPNet-aligned}} \\ 
\cline{4-7} 
                         & & & Annualised IRR & SR & Annualised IRR & SR \\ 
                         \hline
\multirow{6}{*}{NASDAQ}  & GCN                                  & Pairwise Graph    & 0.1402 & 0.4806 & 0.2084 (+0.068) & 0.6891 (+0.209) \\
                         & RSR-I~\cite{feng2019temporal}        & Pairwise Graph    & 0.3919 & 1.3042 & 0.4967 (+0.105) & 1.6365 (+0.332) \\
                         & RT-GCN~\cite{zheng2023relational}    & Pairwise Graph    & 0.3642 & 1.5687 & 0.4717 (+0.110) & 2.2003 (+0.632) \\
                         & Hypergraph Conv                      & Hypergraph        & 0.2064 & 0.6936 & 0.3108 (+0.104) & 1.0288 (+0.335) \\
                         & STHAN-SR \cite{sawhney2021stock}     & Hypergraph        & 0.2382 & 0.7876 & 0.3034 (+0.065) & 0.9866 (+0.199) \\
                         & CI-STHPAN~\cite{xia2024ci}           & Hypergraph        & 0.4515 & 1.6599 & 0.6319 (+0.180) & 2.1190 (+0.459) \\ \hline
\multirow{6}{*}{NYSE}    & GCN                                  & Pairwise Graph    & 0.1620 & 0.4116 & 0.1865 (+0.025) & 0.4502 (+0.039) \\
                         & RSR-I~\cite{feng2019temporal}        & Pairwise Graph    & 0.1651 & 0.4393 & 0.4798 (+0.315) & 0.7128 (+0.274) \\
                         & RT-GCN~\cite{zheng2023relational}    & Pairwise Graph    & 0.1326 & 1.1135 & 0.2250 (+0.092) & 1.5555 (+0.442) \\
                         & Hypergraph Conv                      & Hypergraph        & 0.1326 & 0.3367 & 0.2171 (+0.085) & 0.5760 (+0.239) \\
                         & STHAN-SR~\cite{sawhney2021stock}     & Hypergraph        & 0.0627 & 0.1686 & 0.2157 (+0.153) & 0.6651 (+0.497) \\
                         & CI-STHPAN~\cite{xia2024ci}           & Hypergraph        & 0.1915 & 0.7141 & 0.3195 (+0.128) & 0.8218 (+0.108) \\ 
                         \hline
\end{tabular}%
}
\caption{Profitability improvements and generalisation ability of GAPNet on different backbone model architectures with a recent advanced predefined graph structure (DTW Graph). \textit{Basic with DTW Graph} stands for the two-step paradigm using DTW graph construction ($k=20$), and \textit{GAPNet-aligned} is for our proposed end-to-end paradigm. The values in the brackets denote the absolute improvement in performance achieved by GAPNet. Results are averaged over 5 independent runs (p\textless{}0.01). }
\label{tab:profitability}
\end{table*}

\subsection{Evaluation Metrics}
According to the studies~\cite{feng2019temporal,sawhney2021stock,xia2024ci}, we conduct experiments on a buy-hold-sell trading strategy. 
Specifically, at each trading day $t$, the model predicts the returns of all individual stocks for the next day $t+1$, and selects the top $k$ stocks $S_k$ to hold based on their rank. 
These stocks are traded at the close price of $t+1$, and the daily return is calculated as $R_t = \sum_{i \in S_k}\frac{c_i^{t+1}-c_i^t}{c_i^t}$, where $c_i^t$ denotes the close price of stock $i$ on day $t$. 
We backtest on the entire test set and report the annualised cumulative return (\textbf{annualised IRR}), and the Sharpe Ratio (\textbf{SR}). We also use the information coefficient (\textbf{IC}) and the information coefficient information ratio (\textbf{ICIR}) to further evaluate the predictive accuracy and stability.  

\begin{table*}[]
\centering
\resizebox{\textwidth}{!}{%
\begin{tabular}{lcccc|cccc}
\hline
\multicolumn{1}{c}{\multirow{2}{*}{Graph Information}} & \multicolumn{4}{c|}{CI-STHPAN}                                        & \multicolumn{4}{c}{RT-GCN}                                            \\
\multicolumn{1}{c}{}                                   & Annualised IRR  & SR              & IC              & ICIR            & Annualised IRR  & SR              & IC              & ICIR            \\ \hline
Industry                                               & 0.3121          & 1.2344          & -0.0003         & -0.0050         & 0.2085          & 1.4745          & -0.0016         & -0.0488         \\
Wiki                                                   & 0.4130          & 1.6502          & -0.0004         & -0.0062         & \underline{0.4364}          & \underline{1.9678}          & \underline{0.0055}          & 0.1149          \\
Industry+Wiki                                          & 0.4445          & 1.7319          & -0.0003         & -0.0039         & 0.1049          & 0.3351          & 0.0019          & 0.0470          \\
Industry+DTW                                           & 0.3970          & 1.5729          & -0.0004         & -0.0056         & 0.2516          & 1.4021          & 0.0050          & 0.1374          \\
Wiki+DTW                                               & 0.2623          & 1.0348          & -0.0004         & -0.0053         & 0.2869          & 1.2515          & 0.0029          & 0.0551          \\
DTW\_K=20                                              & \underline{0.4515}          & 1.6599          & \underline{0.0015}          & \underline{0.0178}          & 0.3642          & 1.5687          & 0.0035          & 0.0595          \\
DTW\_K=15                                              & 0.3380          & 1.1970          & -0.0009         & -0.0122         & 0.2673          & 1.1252          & -0.0002         & -0.0034         \\
DTW\_K=10                                              & 0.3172          & 1.2643          & -0.0005         & -0.0069         & 0.0477          & 0.2296          & -0.0005         & -0.0098         \\
DTW\_K=5                                               & 0.3282          & 1.3057          & -0.0003         & -0.0041         & 0.2631          & 1.5231          & 0.0036          & \textbf{0.1618}          \\
DTW\_K=1                                               & 0.4490          & \underline{1.7709}          & -0.0003         & -0.0042         & 0.1280          & 0.8902          & -0.0007         & -0.0267         \\ 
All                                                    & 0.3301          & 1.3155          & -0.0006         & -0.0088         & 0.3021          & 1.7824          & 0.0055          & 0.1605          \\ \hline
\textbf{GAPNet-aligned} & \textbf{0.6319} & \textbf{2.1190} & \textbf{0.0052} & \textbf{0.0597} & \textbf{0.4741} & \textbf{2.2003} & \textbf{0.0068} & \underline{0.1613} \\ \hline
\end{tabular}%
}
\caption{The impact of different graph information on CI-STHPAN and RT-GCN. \textit{Industry} and \textit{Wiki} represent predefined domain knowledge graphs. \textit{DTW\_K} stands for the dynamic relation between $k$ stocks based on the similarity, and \textit{All} is for merging all predefined graphs presented above. Results are averaged over 5 independent runs (p\textless{}0.01).}
\label{tab:relations}
\end{table*}

\subsection{Overall Performance}
In order to evaluate the effectiveness and generalisability of the proposed GAPNet, we conduct extensive experiments on two fundamental GNN architectures, i.e. GCN and hypergraph convolution, as well as four state-of-the-art (SOTA) stock ranking models, which are RSR-I, STHAN-SR, RT-GCN, and CI-STHPAN\footnote{Note that the CI-STHPAN model is trained under supervised learning in our experiments.}. 
In consideration of DGDNN, which frames the prediction of next-day stock trends as a binary temporal node-classification problem~\cite{you2024dgdnn}, while the focus of our study is on the task of stock ranking and selection under portfolio-oriented objectives. The primary focus of DGDNN is classification metrics, whereas the present study assesses ranking and investment performance. In order to ensure a task-aligned comparison, the experiment adopts only DGDNN's dynamic graph construction, i.e. entropy-driven edge generation.
Each model is trained under the traditional two-step paradigm with a predefined graph, and under our end-to-end paradigm, aligned by GAPNet. 
The two variants are designated as the \textit{basic} version and the \textit{GAPNet-aligned} version, respectively. 

In Table \ref{tab:profitability}, GAPNet-aligned versions consistently outperform their basic version across all backbone models on both NASDAQ and NYSE datasets, with significant improvements in annualised IRR and SR ($p < 0.01$). 
The result indicates that involving GAPNet for alignment is necessary and effective for enhancing the backbone models' performance on the downstream task. 
The most significant enhancements of the annualised IRR are observed in the NASDAQ with CI-STHPAN ($+0.180$) and in the NYSE with RSR-I ($+0.315$). The enhancement of SR is most pronounced in the NASDAQ with RT-GCN ($+0.632$) and in the NYSE with STHAN-SR ($+0.497$). 

To independently test the effect of different predefined graph structures on GNN models, we conduct experiments using RT-GCN (pairwise graph-based) and CI-STHPAN (hypergraph-based) on the NASDAQ dataset, with various predefined graph structures (see~\cite{xia2024ci} for graph construction details). The results are reported in Table~\ref{tab:relations}. Consistent findings can be observed on the NYSE dataset, shown in the Appendix. 
The result shows that GNN models' performance is highly sensitive to the choice of graph structure. RT-GCN, in particular, exhibits considerable performance fluctuation across different graph configurations, with annualised IRR ranging from $0.0477$ to $0.4364$, and with SR from $0.2296$ to $1.9678$. This variability is indicative of the non-alignment limitation of predefined graph structures, i.e. \textit{any single predefined graph structure is inherently biased, and there exists no universal guideline for which relations to include or discard}. 
As a result, models built on such graphs inevitably suffer from limited generalisability. The findings demonstrate that the GAPNet-aligned graph consistently outperforms all predefined graphs. 
Furthermore, the GAPNet-aligned graph demonstrates significant improvements in IC and ICIR, reflecting a stronger alignment between predicted and realized returns and a more stable predictive performance over time. Overall, the proposed GAPNet reduces bias by aligning predefined graph structures with downstream tasks in an end-to-end manner, which further promotes a generalised application of all backbone GNN models.

\begin{table}[t]
\centering
\begin{tabular}{cccc|cc}
\hline
GAPNet & \multirow{2}{*}{Initialization} & \multicolumn{2}{c|}{Basic}   & \multicolumn{2}{c}{\textbf{GAPNet-aligned}} \\
Component                           &                                 & Ann.\ IRR      & SR             & Ann.\ IRR      & SR           \\ \hline
SPL+TPL                        & Industry                        & 0.3121         & 1.2344         & 0.5649         & 1.9979       \\
SPL+TPL                        & Wiki                            & 0.4130         & 1.6502         & 0.5265         & 1.8604       \\
SPL+TPL                        & Industry+Wiki                   & 0.4445         & 1.7319         & 0.5906         & 2.0829       \\
SPL+TPL                        & Industry+DTW                    & 0.3970         & 1.5729         & 0.5365         & 1.9025       \\
SPL+TPL                        & Wiki+DTW                        & 0.2623         & 1.0348         & 0.5961         & 2.1196       \\
SPL+TPL                        & \texttt{DTW\_K=20}              & 0.4515         & 1.6599         & 0.6319         & 2.1190       \\
SPL+TPL                        & \texttt{DTW\_K=15}              & 0.3380         & 1.1970         & 0.5422         & 1.9248       \\
SPL+TPL                        & \texttt{DTW\_K=10}              & 0.3172         & 1.2643         & 0.5296         & 1.8833       \\
SPL+TPL                        & \texttt{DTW\_K=5}               & 0.3282         & 1.3057         & 0.5142         & 1.8201       \\
SPL+TPL                        & \texttt{DTW\_K=1}               & 0.4490         & 1.7709         & 0.5621         & 1.9658       \\
SPL+TPL                        & All                             & 0.3301         & 1.3155         & 0.5535         & 1.9590       \\
SPL+TPL                        & Random                          & -              & -              & 0.4876         & 1.8654       \\
w.o. TPL                       & -                               & -              & -              & 0.4542         & 1.7188       \\ \hline
\end{tabular}%
\caption{Ablation study on GAPNet's components and initialization. CI-STHPAN is used as the backbone model. \textit{Basic} represents the two-step paradigm. \textit{GAPNet-aligned} represents the end-to-end paradigm. \textit{Ann.\ IRR} denotes annualized IRR.}
\label{tab:ablation}
\end{table}

\subsection{Ablation Study}
Table \ref{tab:ablation} demonstrates the ablation results on the NASDAQ dataset to assess the contribution of each GAPNet component and graph initialisation. 
Consistent results can be observed on the NYSE dataset, provided in the Appendix. 
We focus on two aspects: 
\begin{enumerate}
    \item the effect of different initialisation strategies (predefined graph construction) for GAPNet.
    \item the impact of removing TPL entirely (denoted as \textit{w.o. TPL}, which GAPNet relying solely on the SPL to capture short-term dependencies. 
\end{enumerate}

\subsubsection{Initialisation Sensitivity}
Firstly, when initialised with different predefined graphs, GAPNet consistently achieves powerful performance with reduced variation in both metrics, demonstrating robustness to graph initialisation and the ability to generalise across diverse prior graph structures. Secondly, even in scenarios where TPL is initialised at random without any prior relational knowledge, GAPNet remains highly competitive, exhibiting only a slight performance drop. This finding indicates that there is less loss in terms of GAPNet's ability to effectively learn dynamic relational structures in the absence of structural priors.

\subsubsection{Effect of SPL and TPL}
\textit{w.o. TPL} leads to a non-ignorable performance drop (IRR = $0.4542$, SR = $1.7188$) compared to the full version of GAPNet with random initialisation (IRR = $0.4876$, SR = $1.8654$), providing evidence that TPL is valuable in integrating short-term and long-term dependencies. Notably, \textit{w.o. TPL} still performs comparably to the best \textit{basic} version, which indicats that a GAPNet with SPL alone is approximately as effective as a predefined graph. These results highlight the independent effects of SPL and TPL, and the benefit of integrating both within the full GAPNet architecture.




\begin{figure}[t]
\centering

\begin{subfigure}[t]{0.48\linewidth}
\centering
\includegraphics[width=\linewidth]{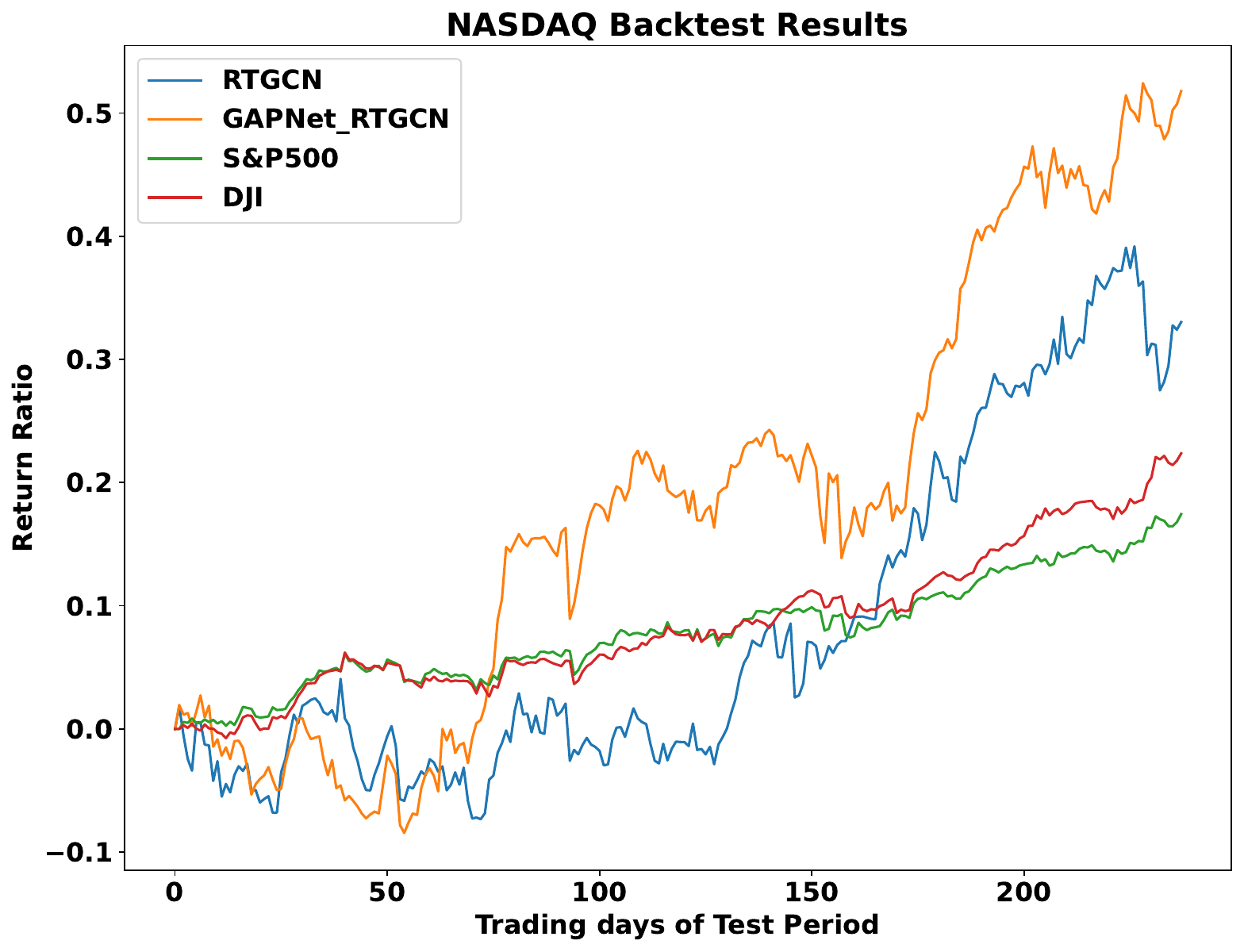}
\caption{RTGCN}
\label{fig:nasdaq_rtgcn}
\end{subfigure}
\hfill
\begin{subfigure}[t]{0.48\linewidth}
\centering
\includegraphics[width=\linewidth]{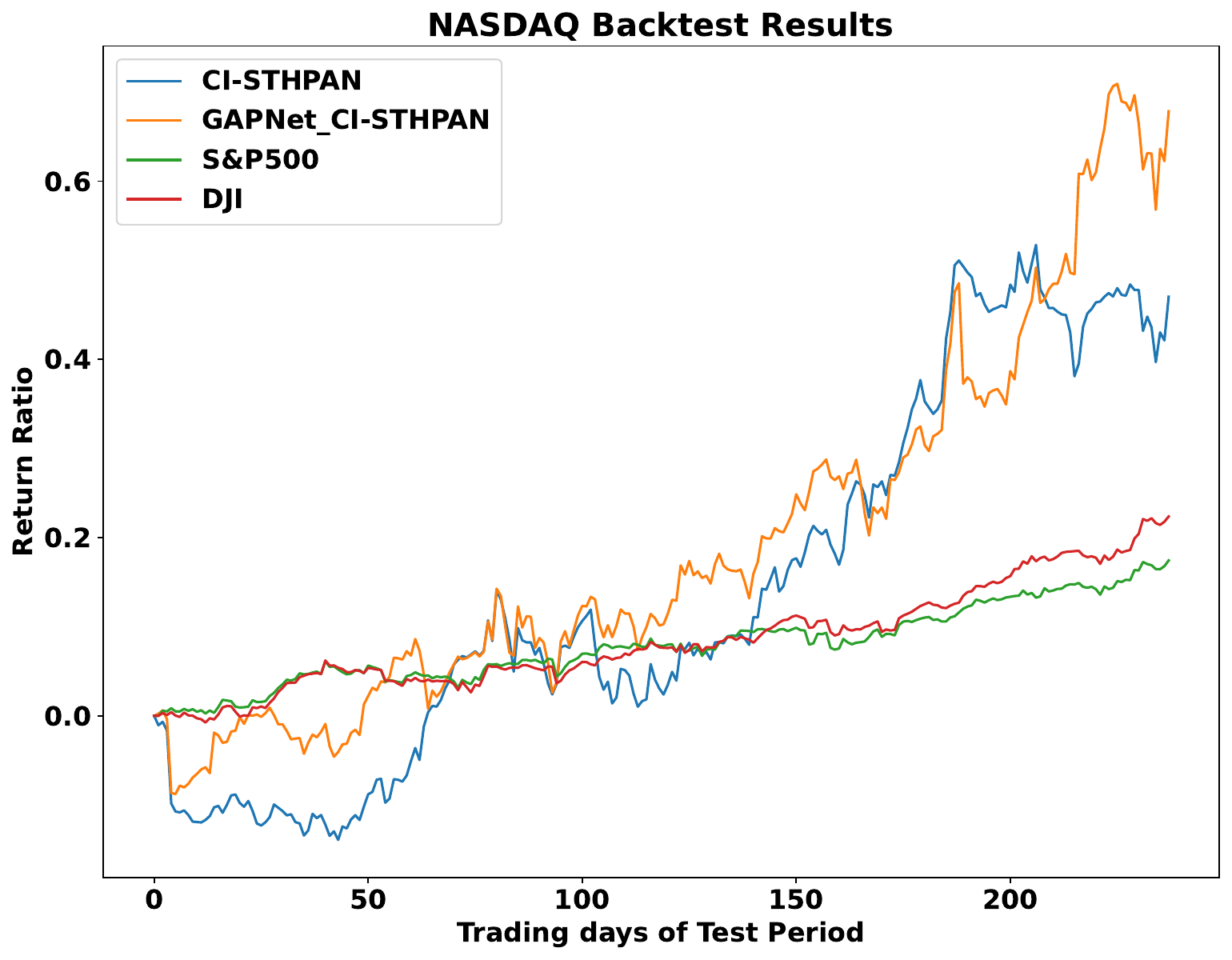}
\caption{CI-STHPAN}
\label{fig:nasdaq_ci}
\end{subfigure}

\caption{Backtesting (top-5) results on NASDAQ dataset with/without GAPNet.}
\label{fig:nasdaq_backtesting}
\end{figure}ECML

\subsection{Backtesting Results}
Figures~\ref{fig:nasdaq_rtgcn} and~\ref{fig:nasdaq_ci} present the simulated backtesting results of two backbone models, RT-GCN and CI-STHPAN, on the NASDAQ testing period. To benchmark the model performance against the market, we include the \textit{Dow Jones Industrial Average (DJI)} and \textit{S\&P 500} indices as reference baselines. 
We adopt an equal-weight investment strategy with an initial capital sum of $10000$. Since the dataset is updated on a daily basis, the portfolio rebalancing frequency is also set to daily. 
Specifically, at the termination of each trading day $t$, the model ranks all candidate stocks and selects the top-$k$ stocks for portfolio construction. 
The selected stocks are equally weighted and held for one day; they are liquidated at the end of $t+1$, and a new portfolio is reconstructed based on the next day’s predictions. To simplify the experimental setup, transaction costs are ignored.

As shown in Figures~\ref{fig:nasdaq_rtgcn} and~\ref{fig:nasdaq_ci}, incorporating GAPNet consistently enhances the investment performance of both backbone models, outperforming their standalone versions and the market benchmarks. For RT-GCN, the cumulative return (CR) and the Sharpe ratio (SR) are 0.33 and 1.46, respectively, while RT-GCN + GAPNet achieves CR = 0.52 and SR = 2.11. Similarly, CI-STHPAN attains CR = 0.47 and SR = 1.75, whereas CI-STHPAN + GAPNet improves these metrics to CR = 0.68 and SR = 2.12. 
These results are consistent with the quantitative findings reported in Table~\ref{tab:profitability}, confirming that integrating GAPNet effectively enhances the backbone models’ downstream performance in stock selection tasks. 
The results of backtesting conducted on the NYSE demonstrate consistent trends, and the details are provided in the Appendix.

\subsection{Computational Cost}
Figure \ref{fig:cost} provides a graphical representation of the training time and GPU memory usage for six backbone models. To assess the scalability of the GAPNet, a comparison was made of resource usage when each backbone is trained under the two-step paradigm versus under the end-to-end paradigm. 
It was observed that the overhead incurred by GAPNet depends on the underlying graph representation. The impact on backbone models based on standard pairwise graphs (e.g., GCN, RSR-I, RT-GCN) in terms of additional training time and memory usage is negligible. 
In contrast, when GAPNet is employed in conjunction with hypergraph-based models (e.g., HGCN, STHAN-SR, CI-STHPAN), there is an observed increase in resource consumption. 
This is primarily attributed to the denser structure of hypergraphs and the more computationally intensive transformation from dense adjacency matrices to hypergraph incidence representations. 
Overall, the additional computational cost is acceptable in comparison to the predefined graph construction and the end-to-end GAPNet-aligned graph with backbone models, which indicates that GAPNet is a feasible option and should be preferentially employed in most practical scenarios.

\begin{figure}[]
\centering
\includegraphics[width=\linewidth]{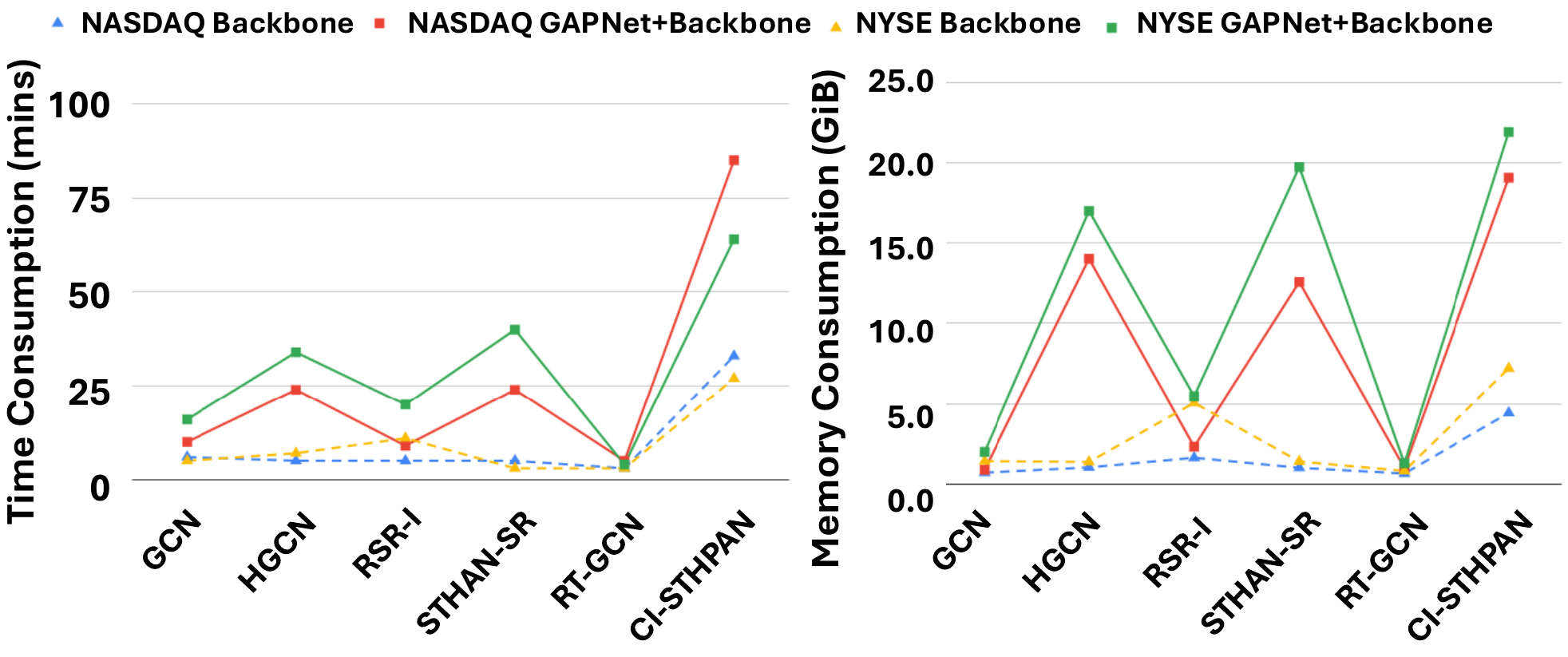} 
\caption{The comparsion of computational cost for six backbone models. Left: training time; Right: memory usage.}
\label{fig:cost}
\end{figure}

\section{Conclusion}
In this paper, we propose a novel end-to-end training paradigm for GNN-based stock ranking task, and GAPNet, a Graph Adaptation Plug-in Network that adaptively aligns predefined graphs for a given downstream task, and improves GNN performance on the task. 
GAPNet addresses the non-alignment and poor generalisation issues that are commonly suffered by predefined graphs. Through extensive experiments on NASDAQ and NYSE, we demonstrate the effectiveness and generalisability of GAPNet and the paradigm across both pairwise graph-based and hypergraph-based backbone models is demonstrated, highlighting its potential in relational modeling for financial forecasting. 
From a design perspective, the proposed paradigm and GAPNet are readily adaptable to any downstream tasks in the same end-to-end training paradigm. This finding provides a foundation for future research, including the potential application of our paradigm and GAPNet to a wide-ranging set of financial forecasting tasks. It is also noteworthy that the proposed paradigm and GAPNet are conceptually independent: GAPNet is merely one implementation of the end-to-end training paradigm. In future studies, alternative designs of end-to-end training paradigm could be explored to further enhance flexibility and performance.


%
%
%
\bibliographystyle{splncs04}
\bibliography{ref}

@inproceedings{chen2018incorporating,
  title={Incorporating corporation relationship via graph convolutional neural networks for stock price prediction},
  author={Chen, Yingmei and Wei, Zhongyu and Huang, Xuanjing},
  booktitle={Proceedings of the 27th ACM international conference on information and knowledge management},
  pages={1655--1658},
  year={2018}
}

@article{chen2020iterative,
  title={Iterative deep graph learning for graph neural networks: Better and robust node embeddings},
  author={Chen, Yu and Wu, Lingfei and Zaki, Mohammed},
  journal={Advances in neural information processing systems},
  volume={33},
  pages={19314--19326},
  year={2020}
}

@inproceedings{cheng2021modeling,
  title={Modeling the momentum spillover effect for stock prediction via attribute-driven graph attention networks},
  author={Cheng, Rui and Li, Qing},
  booktitle={Proceedings of the AAAI conference on artificial intelligence},
  volume={35},
  number={1},
  pages={55--62},
  year={2021}
}

@article{cui2023temporal,
  title={Temporal-relational hypergraph tri-attention networks for stock trend prediction},
  author={Cui, Chaoran and Li, Xiaojie and Zhang, Chunyun and Guan, Weili and Wang, Meng},
  journal={Pattern recognition},
  volume={143},
  pages={109759},
  year={2023},
  publisher={Elsevier}
}

@inproceedings{duan2025factorgcl,
  title={FactorGCL: A Hypergraph-Based Factor Model with Temporal Residual Contrastive Learning for Stock Returns Prediction},
  author={Duan, Yitong and Wang, Weiran and Li, Jian},
  booktitle={Proceedings of the AAAI Conference on Artificial Intelligence},
  volume={39},
  number={1},
  pages={173--181},
  year={2025}
}

@inproceedings{dong2024fnspid,
  title={Fnspid: A comprehensive financial news dataset in time series},
  author={Dong, Zihan and Fan, Xinyu and Peng, Zhiyuan},
  booktitle={Proceedings of the 30th ACM SIGKDD Conference on Knowledge Discovery and Data Mining},
  pages={4918--4927},
  year={2024}
}

@article{feng2018enhancing,
  title={Enhancing stock movement prediction with adversarial training},
  author={Feng, Fuli and Chen, Huimin and He, Xiangnan and Ding, Ji and Sun, Maosong and Chua, Tat-Seng},
  journal={arXiv preprint arXiv:1810.09936},
  year={2018}
}

@article{feng2019temporal,
  title={Temporal relational ranking for stock prediction},
  author={Feng, Fuli and He, Xiangnan and Wang, Xiang and Luo, Cheng and Liu, Yiqun and Chua, Tat-Seng},
  journal={ACM Transactions on Information Systems (TOIS)},
  volume={37},
  number={2},
  pages={1--30},
  year={2019},
  publisher={ACM New York, NY, USA}
}

@inproceedings{franceschi2019learning,
  title={Learning discrete structures for graph neural networks},
  author={Franceschi, Luca and Niepert, Mathias and Pontil, Massimiliano and He, Xiao},
  booktitle={International conference on machine learning},
  pages={1972--1982},
  year={2019},
  organization={PMLR}
}

@article{hochreiter1997long,
  title={Long short-term memory},
  author={Hochreiter, Sepp and Schmidhuber, J{\"u}rgen},
  journal={Neural computation},
  volume={9},
  number={8},
  pages={1735--1780},
  year={1997},
  publisher={MIT press}
}

@inproceedings{jiang2019semi,
  title={Semi-supervised learning with graph learning-convolutional networks},
  author={Jiang, Bo and Zhang, Ziyan and Lin, Doudou and Tang, Jin and Luo, Bin},
  booktitle={Proceedings of the IEEE/CVF conference on computer vision and pattern recognition},
  pages={11313--11320},
  year={2019}
}

@article{kingma2014adam,
  title={Adam: A method for stochastic optimization},
  author={Kingma, Diederik P and Ba, Jimmy},
  journal={arXiv preprint arXiv:1412.6980},
  year={2014}
}

@article{kipf2016semi,
  title={Semi-supervised classification with graph convolutional networks},
  author={Kipf, Thomas N and Welling, Max},
  journal={arXiv preprint arXiv:1609.02907},
  year={2016}
}

@inproceedings{li2021modeling,
  title={Modeling the stock relation with graph network for overnight stock movement prediction},
  author={Li, Wei and Bao, Ruihan and Harimoto, Keiko and Chen, Deli and Xu, Jingjing and Su, Qi},
  booktitle={Proceedings of the twenty-ninth international conference on international joint conferences on artificial intelligence},
  pages={4541--4547},
  year={2021}
}

@article{muhammad2023transformer,
  title={Transformer-based deep learning model for stock price prediction: A case study on Bangladesh stock market},
  author={Muhammad, Tashreef and Aftab, Anika Bintee and Ibrahim, Muhammad and Ahsan, Md Mainul and Muhu, Maishameem Meherin and Khan, Shahidul Islam and Alam, Mohammad Shafiul},
  journal={International Journal of Computational Intelligence and Applications},
  volume={22},
  number={03},
  pages={2350013},
  year={2023},
  publisher={World Scientific}
}

@inproceedings{niu2024evaluating,
  title={Evaluating Financial Relational Graphs: Interpretation Before Prediction},
  author={Niu, Yingjie and Lu, Lanxin and Dolphin, Rian and Poti, Valerio and Dong, Ruihai},
  booktitle={Proceedings of the 5th ACM International Conference on AI in Finance},
  pages={564--572},
  year={2024}
}

@article{niu2025ngat,
  title={NGAT: A Node-level Graph Attention Network for Long-term Stock Prediction},
  author={Niu, Yingjie and Zhao, Mingchuan and Poti, Valerio and Dong, Ruihai},
  journal={arXiv preprint arXiv:2507.02018},
  year={2025}
}

@article{patel2024systematic,
  title={A systematic review on graph neural network-based methods for stock market forecasting},
  author={Patel, Manali and Jariwala, Krupa and Chattopadhyay, Chiranjoy},
  journal={ACM Computing Surveys},
  volume={57},
  number={2},
  pages={1--38},
  year={2024},
  publisher={ACM New York, NY}
}

@inproceedings{qian2024mdgnn,
  title={Mdgnn: Multi-relational dynamic graph neural network for comprehensive and dynamic stock investment prediction},
  author={Qian, Hao and Zhou, Hongting and Zhao, Qian and Chen, Hao and Yao, Hongxiang and Wang, Jingwei and Liu, Ziqi and Yu, Fei and Zhang, Zhiqiang and Zhou, Jun},
  booktitle={Proceedings of the AAAI Conference on Artificial Intelligence},
  volume={38},
  number={13},
  pages={14642--14650},
  year={2024}
}

@inproceedings{sawhney2021stock,
  title={Stock selection via spatiotemporal hypergraph attention network: A learning to rank approach},
  author={Sawhney, Ramit and Agarwal, Shivam and Wadhwa, Arnav and Derr, Tyler and Shah, Rajiv Ratn},
  booktitle={Proceedings of the AAAI Conference on Artificial Intelligence},
  volume={35},
  number={1},
  pages={497--504},
  year={2021}
}

@inproceedings{shang2021discrete,
  title={Discrete Graph Structure Learning for Forecasting Multiple Time Series},
  author={Shang, Chao and Chen, Jie},
  booktitle={Proceedings of International Conference on Learning Representations},
  year={2021}
}

@inproceedings{soun2022accurate,
  title={Accurate stock movement prediction with self-supervised learning from sparse noisy tweets},
  author={Soun, Yejun and Yoo, Jaemin and Cho, Minyong and Jeon, Jihyeong and Kang, U},
  booktitle={2022 IEEE International Conference on Big Data (Big Data)},
  pages={1691--1700},
  year={2022},
  organization={IEEE}
}

@inproceedings{smith2019super,
  title={Super-convergence: Very fast training of neural networks using large learning rates},
  author={Smith, Leslie N and Topin, Nicholay},
  booktitle={Artificial intelligence and machine learning for multi-domain operations applications},
  volume={11006},
  pages={369--386},
  year={2019},
  organization={SPIE}
}

@article{vaswani2017attention,
  title={Attention is all you need},
  author={Vaswani, Ashish and Shazeer, Noam and Parmar, Niki and Uszkoreit, Jakob and Jones, Llion and Gomez, Aidan N and Kaiser, {\L}ukasz and Polosukhin, Illia},
  journal={Advances in neural information processing systems},
  volume={30},
  year={2017}
}

@article{velivckovic2017graph,
  title={Graph attention networks},
  author={Veli{\v{c}}kovi{\'c}, Petar and Cucurull, Guillem and Casanova, Arantxa and Romero, Adriana and Lio, Pietro and Bengio, Yoshua},
  journal={arXiv preprint arXiv:1710.10903},
  year={2017}
}

@inproceedings{wang2022adaptive,
  title={Adaptive Long-Short Pattern Transformer for Stock Investment Selection.},
  author={Wang, Heyuan and Wang, Tengjiao and Li, Shun and Zheng, Jiayi and Guan, Shijie and Chen, Wei},
  booktitle={IJCAI},
  pages={3970--3977},
  year={2022}
}

@inproceedings{wu2018hybrid,
  title={Hybrid deep sequential modeling for social text-driven stock prediction},
  author={Wu, Huizhe and Zhang, Wei and Shen, Weiwei and Wang, Jun},
  booktitle={Proceedings of the 27th ACM international conference on information and knowledge management},
  pages={1627--1630},
  year={2018}
}

@inproceedings{xia2024ci,
  title={CI-STHPAN: pre-trained attention network for stock selection with channel-independent spatio-temporal hypergraph},
  author={Xia, Hongjie and Ao, Huijie and Li, Long and Liu, Yu and Liu, Sen and Ye, Guangnan and Chai, Hongfeng},
  booktitle={Proceedings of the AAAI Conference on Artificial Intelligence},
  volume={38},
  number={8},
  pages={9187--9195},
  year={2024}
}

@inproceedings{xiang2022temporal,
  title={Temporal and heterogeneous graph neural network for financial time series prediction},
  author={Xiang, Sheng and Cheng, Dawei and Shang, Chencheng and Zhang, Ying and Liang, Yuqi},
  booktitle={Proceedings of the 31st ACM international conference on information \& knowledge management},
  pages={3584--3593},
  year={2022}
}

@inproceedings{xu2018stock,
  title={Stock movement prediction from tweets and historical prices},
  author={Xu, Yumo and Cohen, Shay B},
  booktitle={Proceedings of the 56th Annual Meeting of the Association for Computational Linguistics (Volume 1: Long Papers)},
  pages={1970--1979},
  year={2018}
}

@article{xu2022hgnn,
  title={HGNN: Hierarchical graph neural network for predicting the classification of price-limit-hitting stocks},
  author={Xu, Cong and Huang, Huiling and Ying, Xiaoting and Gao, Jianliang and Li, Zhao and Zhang, Peng and Xiao, Jie and Zhang, Jiarun and Luo, Jiangjian},
  journal={Information Sciences},
  volume={607},
  pages={783--798},
  year={2022},
  publisher={Elsevier}
}

@article{you2024dgdnn,
  title={DGDNN: Decoupled graph diffusion neural network for stock movement prediction},
  author={You, Zinuo and Shi, Zijian and Bo, Hongbo and Cartlidge, John and Zhang, Li and Ge, Yan},
  journal={arXiv preprint arXiv:2401.01846},
  year={2024}
}

@inproceedings{you2024multi,
  title={Multi-relational graph diffusion neural network with parallel retention for stock trends classification},
  author={You, Zinuo and Zhang, Pengju and Zheng, Jin and Cartlidge, John},
  booktitle={ICASSP 2024-2024 IEEE International Conference on Acoustics, Speech and Signal Processing (ICASSP)},
  pages={6545--6549},
  year={2024},
  organization={IEEE}
}

@inproceedings{zhang2020spatio,
  title={Spatio-temporal graph structure learning for traffic forecasting},
  author={Zhang, Qi and Chang, Jianlong and Meng, Gaofeng and Xiang, Shiming and Pan, Chunhong},
  booktitle={Proceedings of the AAAI conference on artificial intelligence},
  volume={34},
  number={01},
  pages={1177--1185},
  year={2020}
}

@inproceedings{zheng2023relational,
  title={Relational temporal graph convolutional networks for ranking-based stock prediction},
  author={Zheng, Zetao and Shao, Jie and Zhu, Jia and Shen, Heng Tao},
  booktitle={2023 IEEE 39th International Conference on Data Engineering (ICDE)},
  pages={123--136},
  year={2023},
  organization={IEEE}
}

@article{zhou2006learning,
  title={Learning with hypergraphs: Clustering, classification, and embedding},
  author={Zhou, Dengyong and Huang, Jiayuan and Sch{\"o}lkopf, Bernhard},
  journal={Advances in neural information processing systems},
  volume={19},
  year={2006}
}

\end{document}